\pdfoutput=1

\documentclass[runningheads]{llncs}
\usepackage{graphicx}
\usepackage{subcaption}

\usepackage{tikz}
\usepackage{comment}
\usepackage{amsmath,amssymb} 
\usepackage{color}
\usepackage{multirow}

\usepackage[accsupp]{axessibility}  

\begin{document}
\pagestyle{headings}
\mainmatter
\def\ECCVSubNumber{100}  

\title{HSE-NN Team at the 4th ABAW Competition: Multi-task Emotion Recognition and Learning from Synthetic Images} 

\titlerunning{HSE-NN Team at the 4th ABAW Competition...}
%
\author{Andrey V. Savchenko\orcidID{0000-0001-6196-0564}}
\authorrunning{A.V. Savchenko}
%
\institute{HSE University, Laboratory of Algorithms and Technologies for Network Analysis, Nizhny Novgorod, Russia \\
\email{avsavchenko@hse.ru}}
\maketitle

\begin{abstract}
In this paper, we present the results of the HSE-NN team in the 4th competition on Affective Behavior Analysis in-the-wild (ABAW). The novel multi-task EfficientNet model is trained for simultaneous recognition of facial expressions and prediction of valence and arousal on static photos. The resulting MT-EmotiEffNet extracts visual features that are fed into simple feed-forward neural networks in the multi-task learning challenge. We obtain performance measure 1.3 on the validation set, which is significantly greater when compared to either performance of baseline (0.3) or existing models that are trained only on the s-Aff-Wild2 database. In the learning from synthetic data challenge, the quality of the original synthetic training set is increased by using the super-resolution techniques, such as Real-ESRGAN. Next, the MT-EmotiEffNet is fine-tuned on the new training set. The final prediction is a simple blending ensemble of pre-trained and fine-tuned MT-EmotiEffNets. Our average validation F1 score is 18\% greater than the baseline convolutional neural network. As a result, our team took the first place in the learning from synthetic data challenge and the third place in the multi-task learning challenge.

\end{abstract}

\section{Introduction}
The problem of affective behavior analysis in-the-wild is to understand people's feelings, emotions, and behaviors. Human emotions are typically represented using a small set of basic categories, such as anger or happiness. More advanced representations include a discrete set of Action Units (AUs) from the Facial Action Coding System (FACS) Ekman's model and Russell's continuous encoding of affect in the 2-D space of arousal and valence. The former shows how passive or active an emotional state is, whilst the latter shows how positive or negative it is. Though the emotion of a person may be identified using various signals, such as voice, pronounced utterance, body language, etc., the most accurate results are obtained with facial analytics. 

Due to the high complexity of labeling emotions, existing emotional datasets for facial expression recognition (FER) are small and dirty. As a result, the trained models learn too many features specific to a concrete dataset, which is not practical for in-the-wild settings~\cite{kollias2022abaw}. Indeed, they typically remain not robust to the diversity of environments and video recording conditions, so they can be hardly used in real-world settings with uncontrolled observation conditions. Hence, a lot of attention has been recently brought towards mitigating algorithmic bias in models and, in particular, cross-dataset studies.

This problem has become a focus of many researchers since an appearance of a sequence of the Affective Behavior Analysis in-the-wild (ABAW) challenges that involve different parts of the Aff-Wild~\cite{kollias2019deep,zafeiriou2017aff} and Aff-Wild2~\cite{kollias2017recognition,kollias2019expression} databases. The organizers encourage participants to actively pre-train the models on other datasets by introducing all the new requirements. For example, one of the tasks of the third ABAW competition was the multi-task-learning (MTL) for simultaneous prediction of facial expressions, valence, arousal, and AUs~\cite{kollias2022abaw,kollias2021distribution}. Its winners~\cite{Deng_2022_CVPR} did not use the MTL small static training set (s-Aff-Wild2). Indeed, they proposed the Transformer-based Sign-and-Message Multi-Emotion Net that was trained on a large set of all video frames from the Aff-Wild2. The runner-up proposed the auditory-visual representations~\cite{Jeong_2022_CVPR} that were also trained on initial video files. Similarly, the team that took fourth place in the MTL challenge developed the transformer-based multimodal framework~\cite{Zhang_2022_CVPR} trained on the video frames from the Aff-Wild2 dataset that let them become the winners of FER and AU sub-challenges. 

As a result, in the fourth ABAW competition~\cite{kollias2022abaw4}, such usage of other data from Aff-Wild2 except the small training set is prohibited. It seems that only one successful participant (the third place) of the previous MTL challenge, namely, multi-head EfficientNet~\cite{savchenko2022cvprw}, and the baseline VGGFACE convolutional neural network (CNN)~\cite{kollias2022abaw} satisfy new requirements. Moreover, a new challenge has been introduced that encourages the refinement of the pre-trained models on a set with synthetic faces generated from a small part of the Aff-Wild2 dataset~\cite{kollias2020deep,kollias2020va,kollias2018photorealistic}.

In this paper, we discuss our solution for all tasks from the ABAW4 competition. It is proposed to improve the EfficientNet-based model~\cite{savchenko2022cvprw} by pre-training a model on the AffectNet dataset~\cite{mollahosseini2017affectnet} not only for FER but for additional prediction of valence and arousal. The visual embeddings are extracted from the penultimate layer of the resulting MT-EmotiEffNet, while the valence, arousal, and logits for each emotion are obtained at the output of its last layer. The multi-output feed-forward neural network is trained using the s-Aff-Wild2 database for the MTL challenge. The best validation results are obtained by simple blending~\cite{savchenko2020ad} of its predictions with action unit features from the OpenFace 2 toolkit~\cite{baltrusaitis2018openface}. In the Learning from Synthetic Data (LSD) challenge, we propose to increase the quality of the original synthetic training set by using the super-resolution techniques, such as Real-ESRGAN~\cite{wang2021real}, and fine-tuning the MT-EmotiEffNet on the new training set. The final prediction is a simple blending of scores at the output of pre-trained and fine-tuned MT-EmotiEffNets. The source code for the proposed solutions are made publicly available\footnote{\url{https://github.com/HSE-asavchenko/face-emotion-recognition/blob/main/src/ABAW}}.

This paper is organized as follows. Section~\ref{sec:2} introduces the MT-EmotiEffNet model and the training procedures for both tasks. Experimental results are presented in Section~\ref{sec:3}. Concluding comments are discussed in Section~\ref{sec:4}.
 

\section{Proposed Approach}\label{sec:2}
\subsection{Multi-task learning challenge}\label{subsec:2.1}

The main task of human affective behavior analysis is FER. It is a typical problem of image recognition in which an input facial image $X$ should be associated with one of $C_{EXPR}>1$ categories (classes), such as anger, surprise, etc. There exist several commonly-used expression representations, such as estimation of Valence $V$ and Arousal $A$ (typically, $V, A \in [-1,1]$) and AU detection~\cite{kollias2022abaw}. The latter task is a multi-label classification problem, i.e., prediction of a binary vector $\mathbf{AU}=[AU_1,..., AU_{C_{AU}}]$, where $C_{AU}$ is the total number of AUs, and $AU_i \in \{0,1\}$ is a binary indicator of the presence of the $i$-th AU in the photo.
 
\begin{figure}[t]
\centering
\includegraphics[height=8.7cm]{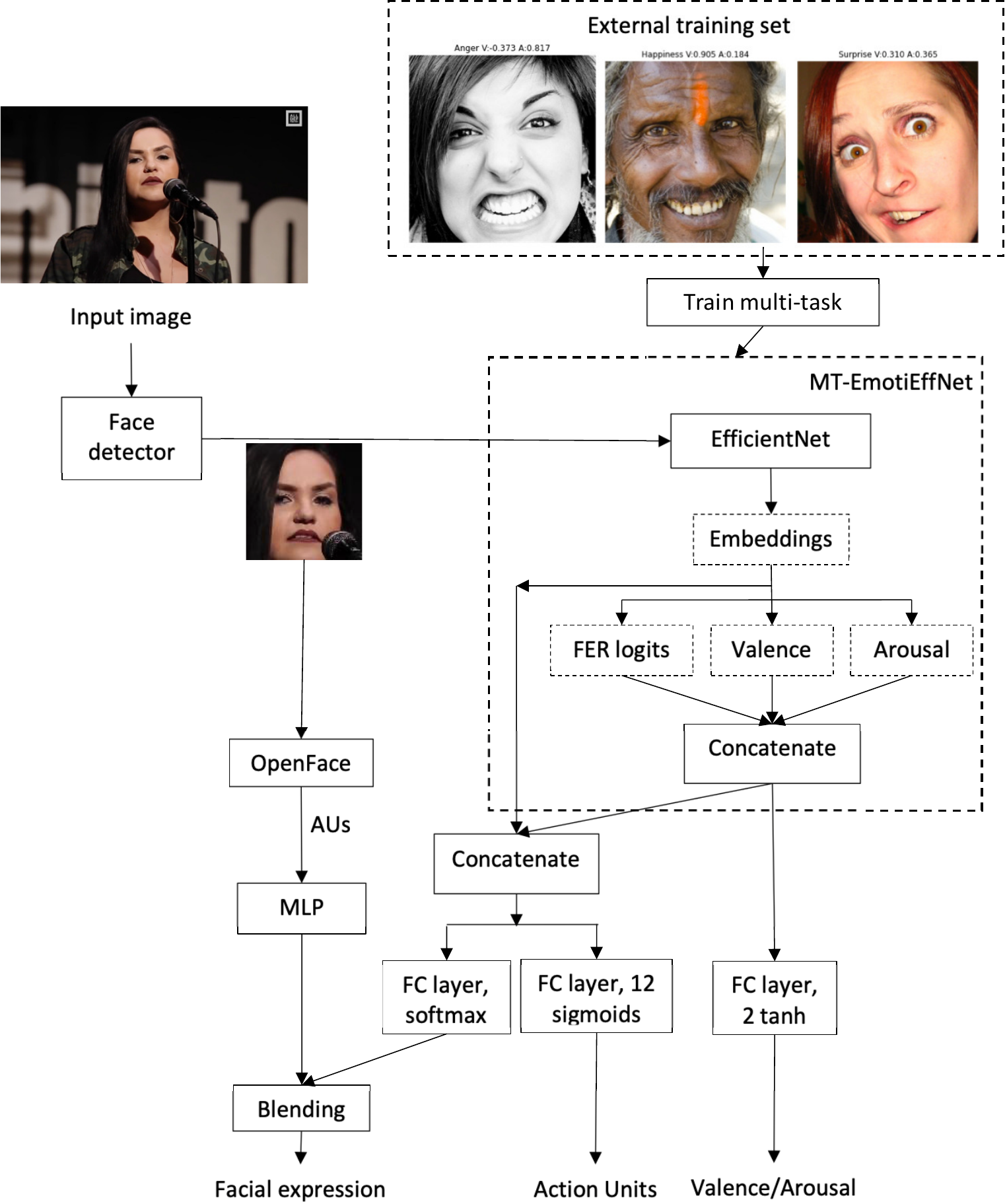}
\caption{Proposed model for the multi-task-learning challenge}
\label{fig:1}
\end{figure}

In this paper, we propose the novel model (Fig.~\ref{fig:1}) for the MTL challenge from ABAW4~\cite{kollias2022abaw}, in which the method from~\cite{savchenko2022cvprw} was modified as follows:
\begin{enumerate}
 \item Emotional feature extractor is implemented with the novel MT-EmotiEffNet model based on EfficientNet-B0 architecture~\cite{tan2019efficientnet}, which was pre-trained not only for FER but for additional valence-arousal estimation.
 \item The valence and arousal for the input facial photo are predicted by using only the output of the last layer of the MT-EmotiEffNet, i.e., they are not concatenated with embeddings extracted at its penultimate layer.
 \item Three heads of the model for facial expressions, Valence-Arousal and AUs, respectively, are trained sequentially, i.e., we do not need additional masks for missing data.
\end{enumerate}

Let us consider the details of this pipeline. Its main part, namely, the MT-EmotiEffNet model, was pre-trained using PyTorch framework  identically to EfficientNet-B0 from~\cite{savchenko2021emotions} on face identification task. The facial regions in the training set are simply cropped by a face detector without margins or face alignment. Next, the last classification layer is replaced by a dense layer with 10 units for valence, arousal, and $C^{(e)}=8$ emotional categories (Neutral, Happy, Sad, Surprise, Fear, Anger, Disgust, Contempt) from the AffectNet dataset~\cite{mollahosseini2017affectnet}, respectively. The loss function is computed as a sum of Concordance Correlation Coefficients (CCCs)~\cite{kollias2021affect} and the weighted categorical cross-entropy~\cite{mollahosseini2017affectnet}:
\begin{multline}
\label{eq:1}
 L(X,y^{(V)},y^{(A)},y^{(e)}) =1-0.5(CCC(z^{(V)},y^{(V)})+CCC(z^{(A)},y^{(A)}))- \\ -\log softmax(z_{y^{(e)}}) \cdot \underset{c \in \{1,...,C^{(e)}\}} \max N_c / N_{y^{(e)}},
\end{multline}
where $X$ is the training image, $y^{(e)} \in \{1,...,C_e\}$ is its emotional class label, $y^{(V)}$ and $y^{(A)}$ are the ground-truth valence and arousal, $N_c$ is the total number of training examples of the $c$-th class, $z_{y^{(e)}}$ is the FER score, i.e., $y^{(e)}$-th output of the last (logits) layer, $z^{(V)}$ and $z^{(A)}$ are the outputs of the last two units in the output layer, and $softmax$ is the softmax activation function.

The imbalanced training set with 287,651 facial photos provided by the authors of the AffectNet~\cite{mollahosseini2017affectnet} was used to train the MT-EmotiEffNet model, while the official balanced set of 4000 images (500 per category) was used for validation. At first, all weights except the new head were frozen, and the model was learned in 3 epochs using the Adam optimizer with a learning rate of 0.001 and SAM (Sharpness-Aware Minimization)~\cite{foret2020sharpness}. Finally, we trained all weights of the model totally of 6 epochs with a lower learning rate (0.0001).

It is important to emphasize that the MT-EmotiEffNet feature extractor is not refined on the s-Aff-Wild2 dataset for the MTL challenge. Thus, every input $X$ and reference $X_n$ image is resized to 224x224 and fed into our CNN. We examine two types of features: (1) facial image embeddings (output of the penultimate layer)~\cite{savchenko2021emotions}; and (2) logits (predictions of emotional unnormalized probabilities at the output of the last layer). The outputs of penultimate layer~\cite{savchenko2021emotions} are stored in the $D=1280$-dimensional embeddings $\mathbf{x}$ and $\mathbf{x}_n$, respectively. The concatenation of 8 FER logits, Valence, and Arousal at the output of the model are stored in the 10-dimensional logits $\mathbf{l}$ and $\mathbf{l}_n$. We experimentally noticed that the valence and arousal for the MTL challenge are better predicted by using the logits only, while the facial expression and AUs are more accurately detected if the embeddings $\mathbf{x}$ and logits $\mathbf{l}$ are concatenated.

The remaining part of our neural network (Fig.~\ref{fig:1}) contains three output layers, namely, (1) $C_{EXPR}=8$ units with softmax activation for recognition of one of eight emotions (Neutral, Anger, Disgust, Fear, Happiness, Sadness, Surprise, Other); (2) two neurons with $tanh$ activation functions for Valence-Arousal prediction; and (3) $C_{AU}=12$ output units with sigmoid activation for AU detection. The model was trained using the s-Aff-Wild2 cropped\_aligned set provided by the organizers~\cite{kollias2022abaw}. This set contains $N=142,333$ facial images $\{X_n\}, n \in \{1,...,N\}$, for which the expression $e_n\in \{1,...,C_{EXPR}\}$, $C_{AU}$-dimensional binary vector $\mathbf{AU}_n$ of AUs, and/or valence $V_n$ and arousal $A_n$ are known. Some labels are missed, so only 90,645 emotional labels, 103,316 AU labels, and 103,917 values of Valence-Arousal are available for training. The validation set contains 26,876 facial frames, for which AU and VA are known, but only 15,440 facial expressions are provided. The three heads of the model, i.e., three fully-connected (FC) layers, were trained separately by using the Tensorflow 2 framework and the procedure described in the paper~\cite{savchenko2022cvprw} for the uni-task learning. 

Finally, we examined the possibility to improve the quality by using additional facial features. The OpenFace 2 toolkit~\cite{baltrusaitis2018openface} extracted pose, gaze, eye, and AU features from each image. It was experimentally found that only the latter features are suitable for the FER part of the MTL challenge. Hence, we trained MLP (multi-layered perceptron) that contains an input layer with 35 AUs from the OpenFace, 1 hidden layer with 128 units and ReLU activation, and 1 output layer with $C_{EXPR}$ units and softmax activation. The component-wise weighted sum of the $C_{EXPR}$ outputs of this model and the emotional scores (posterior probabilities) at the output of ``FC layer, softmax" (Fig.~\ref{fig:1}) is computed in the ``Blending" layer, which returns the emotional class that corresponds to the maximal component of this weighted sum. The best weight in a blending is estimated by maximizing the average F1 score of FER on the validation set. 

\subsection{Learning from synthetic data challenge}\label{subsec:2.2}

\begin{figure}[t]
\centering
\includegraphics[height=7.5cm]{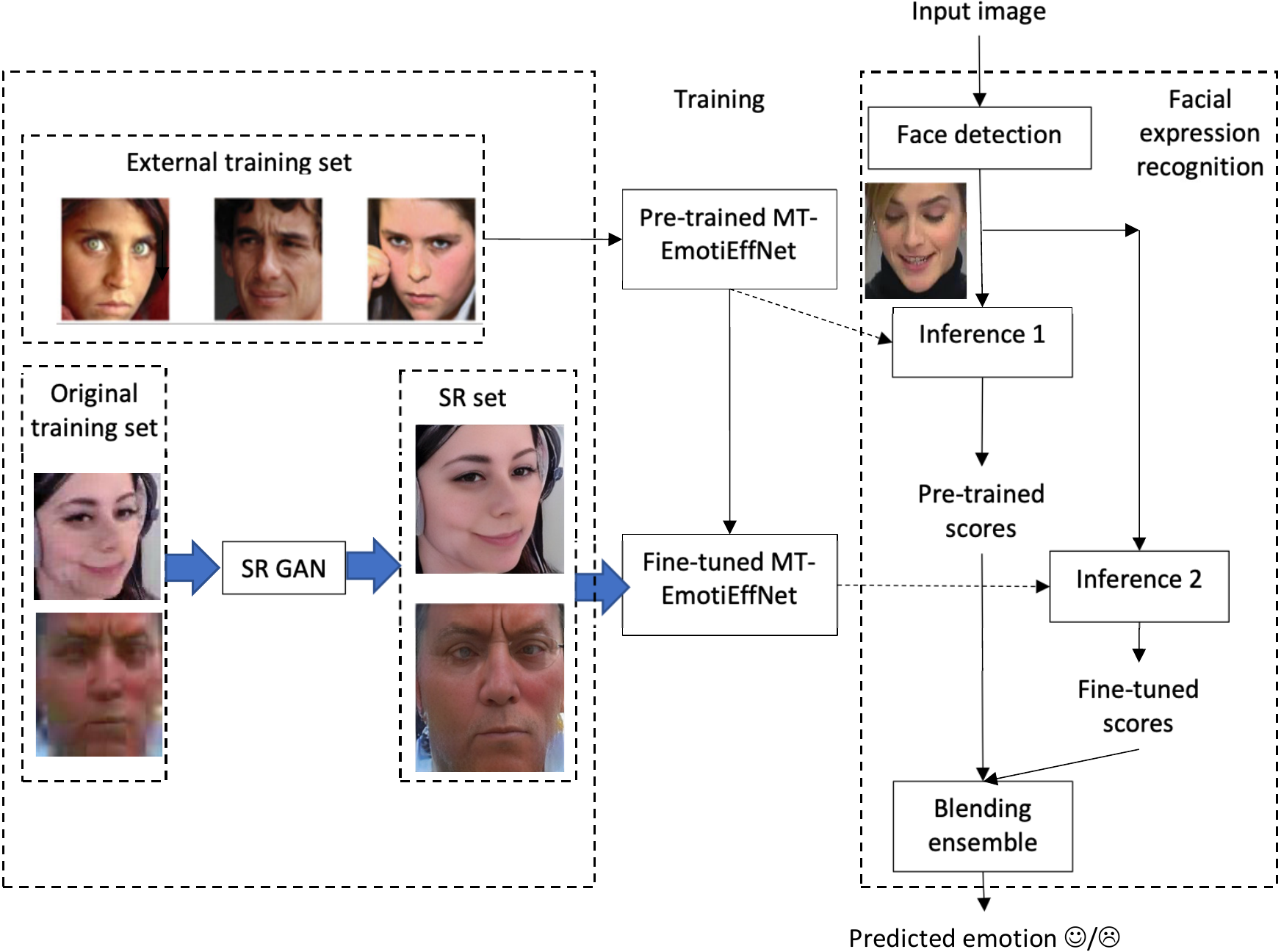}
\caption{Proposed pipeline for learning from synthetic data}
\label{fig:2}
\end{figure}

In the second sub-challenge, it was required to solve the FER task and associate an input facial image with one of $C_{EXPR}=6$ basic expressions (Surprise, Fear, Disgust, Anger, Happiness, Sadness). The task is to refine the pre-trained model by using only information from 277,251 synthetic images that have been generated from some specific frames from the Aff-Wild2 database. The validation set contains 4,670 images for the \textit{same} subjects.

The proposed pipeline (Fig.~\ref{fig:2}) uses the MT-EmotiEffNet from the previous Subsection in a straightforward way. Foremost, we noticed that the quality of provided synthetic facial images with a resolution of 112x112 is too low for our model that was trained on photos with a rather large resolution (224x224). Hence, it was proposed to enhance these images by using contemporary super-resolution (SR) techniques, such as Real-ESRGAN~\cite{wang2021real}. In particular, the pre-trained model RealESRGAN\_x4plus with scale 2 was used for all training images.

Since the required set of 6 expressions is a subset of eight emotions from AffectNet, the simplest solution is to ignore the training set of synthetic images, and predict expression for an image $X$ with the pre-trained MT-EmotiEffNet by using only 6 scores (emotional posterior probabilities) $\mathbf{s}^{(pre)}(X)$ from the last (softmax) layer~\cite{savchenko2021emotions} that is associated with required categories. 

To use the provided synthetic data, we additionally fine-tuned the MT-EmotiEffNet model on the training set at the output of Real-ESRGAN. At first, the output layer was replaced with the new fully-connected layer with $C_{EXPR}$ units, and the weights of the new head were learned during 3 epochs. The remaining weights were frozen. The weighted cross-entropy similar to (\ref{eq:1}) was minimized by the SAM~\cite{foret2020sharpness} with Adam optimizer and learning rate of 0.001. Moreover, we examine the subsequent fine-tuning of all weights on synthetic data during 6 epochs with a learning rate of 0.0001. We can use either model with a new head or completely fine-tuned EfficientNet to predict $C_{EXPR}$-dimensional vector of scores $\mathbf{s}^{(ft)}(X)$ for the input image. To make a final decision, blending of pre-trained and fine-tune models is again applied by computing the weighted sum of scores: $\mathbf{s}(X)=w \cdot \mathbf{s}^{(pre)}(X)+(1-w) \cdot \mathbf{s}^{(ft)}(X)$~\cite{savchenko2020ad}. The final decision is made in favor of the expression that corresponds to the maximal component of the vector $\mathbf{s}(X)$. The best value of the weight hyperparameter $w \in [0,1]$, is estimated by maximizing the average F1 score on the validation set. 

\section{Experimental study}\label{sec:3}

\subsection{FER for static images}
In this subsection, we compare the proposed MT-EmotiEffNet with several existing techniques for static photos from the validation set of AffectNet~\cite{mollahosseini2017affectnet}. The accuracy of FER with 8 emotional categories and CCC and RMSE (root mean squared error) for valence and arousal estimates are summarized in Table~\ref{table:affectnet}.

\setlength{\tabcolsep}{4pt}
\begin{table}[t]
\begin{center}
\caption{Results for the AffectNet validation set (high accuracy and CCC are better, low RMSE is better)}
\label{table:affectnet}
\begin{tabular}{llllll}
\hline\noalign{\smallskip}
 & Facial expressions & \multicolumn{2}{c}{Valence} & \multicolumn{2}{c}{Arousal} \\
Model & 8-class accuracy, \% & RMSE & CCC & RMSE & CCC \\
\noalign{\smallskip}
\hline
\noalign{\smallskip}
AlexNet~\cite{mollahosseini2017affectnet} & 58.0 & 0.394 & 0.541 & 0.402 & 0.450\\
EfficientNet-B0~\cite{savchenko2021emotions} & 61.32 & - & - & - & - \\
SL + SSL inpanting-pl (B0)~\cite{pourmirzaei2021using} & 61.72  & - & - & - & -\\
Distract Your Attention~\cite{wen2021distract} & 62.09  & - & - & - & -\\
EfficientNet-B2~\cite{savchenko2022cvprw} & 63.03  & - & - & - & -\\
\hline
MT-EmotiEffNet & 61.93 & 0.434 & 0.594 & 0.387 & 0.549\\
\hline
\end{tabular}
\end{center}
\end{table}
\setlength{\tabcolsep}{1.4pt}

Though the MT-EmotiEffNet is not the best FER model, it has 0.6\% greater accuracy when compared to a single-task model with identical training settings and architecture (EfficientNet-B0)~\cite{savchenko2021emotions}. One can conclude that taking valence-arousal into account while training the model makes it possible to simultaneously solve multiple affect prediction tasks and extract better emotional features. The RMSE for valence prediction of the AlexNet~\cite{mollahosseini2017affectnet} is slightly better than the RMSE of our model. However, as we optimized the CCC for valence and arousal (\ref{eq:1}), these metric is higher when compared to baseline EfficientNet in all cases. 

\subsection{Multi-task-learning challenge}

In this subsection, the proposed pipeline (Fig.~\ref{fig:1}) for the MTL challenge is examined. At first, we study various features extracted by either known models (OpenFace AUs, EfficientNet-B0~\cite{savchenko2021emotions} ) and our MT-EmotiEffNet. Visual embeddings $\mathbf{x}$ and/or logits $\mathbf{l}$ are fed into MLP with 3 heads. Three training datasets were used, namely, cropped and cropped\_aligned (hereinafter ``aligned) provided by the organizers of this challenge and the aligned faces after Real-ESRGAN~\cite{wang2021real} as we proposed in Subsection~\ref{subsec:2.2}. 

\setlength{\tabcolsep}{4pt}
\begin{table}[t]
\begin{center}
\caption{Ablation study for the MTL challenge}
\label{table:mtl_ablation}
\begin{tabular}{lllllll}
\hline\noalign{\smallskip}
 CNN & Features & Dataset & $P_{MTL}$ & $P_{VA}$ & $P_{EXPR}$ & $P_{AU}$ \\
 \noalign{\smallskip}
\hline
\noalign{\smallskip}
& & cropped & 1.123 & 0.386 & 0.283 & 0.455  \\
 EfficientNet-B0~\cite{savchenko2021emotions} & embeddings + logits& SR & 1.072 &  0.371 & 0.246 & 0.455  \\
& & aligned & 1.148 & 0.396 & 0.290 & 0.462  \\
\hline %
MT-EmotiEffNet & logits & aligned& 1.180 & 0.420 & 0.311 & 0.449 \\
(simultaneous & logits & SR & 1.142 & 0.398 & 0.289 & 0.456  \\
training & embeddings & aligned & 1.190 & 0.415 & 0.308 & 0.467  \\
of heads)& embeddings + logits & aligned & 1.211 & 0.414 & 0.334 & 0.463  \\
\hline %
MT-EmotiEffNet & logits & aligned& 1.157 & 0.443 & 0.260 & 0.454 \\
(separate & embeddings & aligned  & 1.250 & 0.424 & 0.339 & 0.487  \\
training & embeddings + logits & aligned & 1.236 & 0.417 & 0.333 & 0.486  \\
of heads)& our model & aligned & 1.276 & \bf 0.447 & 0.335 & 0.494 \\
\hline
\multicolumn{2}{c}{OpenFace 2 AUs~\cite{baltrusaitis2018openface}} & aligned & 0.900 & 0.242 & 0.256 & 0.402 \\
\hline
\multicolumn{2}{c}{Proposed complete model (Fig.~\ref{fig:1})} & aligned & \bf 1.300 & \bf 0.447 & \bf 0.357 & \bf 0.496\\
\hline
\end{tabular}
\end{center}
\end{table}
\setlength{\tabcolsep}{1.4pt}

The ablation results are presented in Table~\ref{table:mtl_ablation}. Here, we used the performance metrics recommended by the organizers of the challenge, namely, $P_{VA}$ is the average CCC for valence and arousal, $P_{EXPR}$ is the macro-average F1 score for FER, $P_{AU}$ is the macro-average F1 score for AU detection, and $P_{MTL}=P_{VA}+P_{EXPR}+P_{AU}$. The best result in each column is marked in bold. 

The proposed MT-EmotiEffNet has 0.06-0.07 greater overall quality $P_{MTL}$ when compared to EfficientNet-B0 features~\cite{savchenko2021emotions} even if all heads are trained simultaneously as described in the original paper~\cite{savchenko2022cvprw}. If the heads are trained separately without adding masks for missing values, $P_{MTL}$ is increased by 0.02-0.04. Moreover, it was experimentally found that valence and arousal are better predicted by using the logits only. Hence, we developed a multi-head MLP (Fig.~\ref{fig:1}) that feeds logits to the Valence-Arousal head but concatenates logits with embeddings for FER and AU heads. As a result, $P_{VA}$ was increased to 0.447, and the whole model reached $P_{MTL}=1.276$. Finally, we noticed that AU features at the output of OpenFace may be used for rather accurate FER, so that their ensemble with our model is characterized by the best average F1 score for facial expressions $P_{EXPR}=0.357$. It is remarkable that AU features from OpenFace are not well suited for the AU detection task in the MTL challenge, and their blending with our model cannot significantly increase $P_{AU}$.

\setlength{\tabcolsep}{4pt}
\begin{table}
\begin{center}
\caption{Detailed results of the MT-EmotiEffNet for Action Unit detection}
\label{table:best_au}
\begin{tabular}{p{0.18\linewidth}|llllllllllll}
\hline\noalign{\smallskip}
& \multicolumn{12}{c}{Action Unit \#} \\
 & 1  &  2 &  4 &  6 &  7 &  10 &  12 &  15 &  23 &  24 &  25 &  26 \\
\noalign{\smallskip}
\hline
\noalign{\smallskip}
F1 score, thresholds 0.5 & 0.58 & 0.41 & 0.58 & 0.59 & 0.73 & 0.72 & 0.68 & 0.17 & 0.13 & 0.17 & 0.85 & 0.35 \\
\hline
The best thresholds & 0.6 & 0.7 & 0.5 & 0.3 & 0.4 & 0.4 & 0.5 & 0.9 & 0.8 & 0.7 & 0.2 & 0.7 \\
F1 score, the best thresholds & 0.58 & 0.47 & 0.58 & 0.60 & 0.74 & 0.72 & 0.68 & 0.25 & 0.18 & 0.21 & 0.88 & 0.36 \\
\hline
\end{tabular}
\end{center}
\end{table}
\setlength{\tabcolsep}{1.4pt}

\setlength{\tabcolsep}{4pt}
\begin{table}
\begin{center}
\caption{Class-level F1 score of the MT-EmotiEffNet for the FER task}
\label{table:best_expr}
\begin{tabular}{llllllll|l}
\hline\noalign{\smallskip}
Neutral & Anger & Disgust & Fear & Happy & Sad & Surprise & Other  & Total $P_{EXPR}$ \\
\noalign{\smallskip}
\hline
\noalign{\smallskip}
0.3309 & 0.2127 & 0.4307 & 0.2674 & 0.4919 & 0.4534 & 0.1772 & 0.3164 & 0.3351 \\
\hline
\end{tabular}
\end{center}
\end{table}
\setlength{\tabcolsep}{1.4pt}

\setlength{\tabcolsep}{4pt}
\begin{table}
\begin{center}
\caption{Detailed results of the MT-EmotiEffNet for Valence-Arousal prediction}
\label{table:best_va}
\begin{tabular}{lll}
\hline\noalign{\smallskip}
CCC-Valence & CCC-Arousal & Mean CCC $P_{VA}$ \\
\noalign{\smallskip}
\hline
\noalign{\smallskip}
0.4749 & 0.4192 &  0.4471 \\
\hline
\end{tabular}
\end{center}
\end{table}
\setlength{\tabcolsep}{1.4pt}

\setlength{\tabcolsep}{4pt}
\begin{table}
\begin{center}
\caption{Multi-Task-Learning challenge results on the s-Aff-Wild2 validation set}
\label{table:mtl}
\begin{tabular}{llllll}
\hline\noalign{\smallskip}
 Model & AU thresholds & $P_{MTL}$ & $P_{VA}$ & $P_{EXPR}$ & $P_{AU}$ \\
 \noalign{\smallskip}
\hline
\noalign{\smallskip}
VGGFACE Baseline~\cite{kollias2022abaw4} & n/a & 0.30 & - & - & - \\
 \hline
 EfficientNet-B2~\cite{savchenko2022cvprw} & variable & 1.176 & 0.384 & 0.302 & 0.490 \\
\hline %
MT-EmotiEffNet& 0.5 & 1.276 & \multirow{2}{*}{0.447} & \multirow{ 2}{*}{0.335} & 0.494 \\
 & variable & 1.304 & &  & 0.522 \\
\hline %
Proposed complete & 0.5 & 1.300 & \multirow{2}{*}{0.447} & \multirow{ 2}{*}{0.357} & 0.496 \\
model (Fig.~\ref{fig:1}) & variable & 1.326 & &  & 0.522 \\
\hline
\end{tabular}
\end{center}
\end{table}
\setlength{\tabcolsep}{1.4pt}

The detailed results of the proposed model (Fig.~\ref{fig:1}) for AU detection with the additional tuning of thresholds for AU detection, FER, and Valence-Arousal estimation are shown in Table~\ref{table:best_au}, Table~,\ref{table:best_expr} and Table~\ref{table:best_va}, respectively. A comparison of our best model to existing baselines is summarized in Table~\ref{table:mtl}. As one can notice, the usage of the proposed MT-EmotiEffNet significantly increased the performance of valence-arousal prediction and expression recognition. The AUs are also detected with up to 0.03 greater F1 scores. As a result, the  performance measure $P_{MTL}$ of our model is 0.15 and 1.02 points greater when compared to the best model from the third ABAW challenge trained on s-Aff-Wild2 dataset only~\cite{savchenko2022cvprw} and the baseline of the organizers~\cite{kollias2022abaw4}, respectively.

\subsection{Learning from synthetic data challenge}

In this Subsection, our results for the LSD challenge are presented. We examine various EfficientNet-based models that have been pre-trained on AffectNet and fine-tuned on both an original set of synthetic images provided by the organizers and its enhancement by using SR techniques~\cite{wang2021real}.  The official validation set of the organizers without SR was used in all experiments to make the metrics directly comparable. The macro-averaged F1 scores of individual models are presented in Table~\ref{table:synth_ablation}.

\setlength{\tabcolsep}{4pt}
\begin{table}
\begin{center}
\caption{F1 score of single models for the learning from synthetic data challenge}
\label{table:synth_ablation}
\begin{tabular}{lllllll}
\hline\noalign{\smallskip}
  & Pre-& \multicolumn{2}{l}{Fine-tuned (orig)}& \multicolumn{2}{l}{Fine-tuned (SR)}\\
 CNN & trained &  head only  & all weights & head only  & all weights\\
 \noalign{\smallskip}
\hline
\noalign{\smallskip}
 EfficientNet-B0~\cite{savchenko2021emotions} &  0.5972 & 0.5824 & \bf 0.6490 & 0.5994 & 0.6718 \\
 EfficientNet-B2~\cite{savchenko2022cvprw} & 0.5206 & 0.5989 & 0.6412 & 0.6005& 0.6544 \\ 
 MT-EmotiEffNet & \bf 0.6094 & \bf 0.6111 & 0.6324 & \bf 0.6198 & \bf 0.6729\\
\hline
\end{tabular}
\end{center}
\end{table}
\setlength{\tabcolsep}{1.4pt}

Here, even the best pre-trained model is 9-10\% more accurate than the baseline of the organizers (ImageNet ResNet-50) that reached an F1 score of 0.5 on the validation set~\cite{kollias2022abaw4}. As a result, if we replace the last classification layer with a new one and refine the model by training only the weights of the new head, the resulting F1 score will not be increased significantly. It is worth noting that the MT-EmotiEffNet is preferable to other emotional models in all these cases. Thus, our multi-task pre-training seems to provide better emotional features when compared to existing models. As was expected, the fine-tuning of all weights lead to approximately the same results for the same EfficientNet-B0 architecture. It is important to emphasize that the fine-tuning of the model on the training set after Real-ESRGAN leads to a 3-4\% greater F1 score after tuning the whole model. Even if only a new head is learned, performance is increased by 1\% when SR is applied.

The ablation study of the proposed ensemble (Fig.~\ref{fig:2}) is provided in Table~\ref{table:synth_ensemble}. We noticed that as the synthetic data have been generated from subjects of the validation set, but not of the test set, the baseline's performance on the validation and test sets (real data from Aff-Wild2) is equal to 0.50 and 0.30, respectively~\cite{kollias2022abaw4}. Thus, the high accuracy on the validation set does not necessarily lead to the high testing quality. Hence, we estimated the F1 score not only on the official validation set from this sub-challenge but also on the validation set from the MTL competition. The frames with Neutral and Other expressions were removed to obtain a new validation set of 8,953 images with 6 expressions. 

\setlength{\tabcolsep}{4pt}
\begin{table}
\begin{center}
\caption{F1 score of ensemble models for the learning from synthetic data challenge, CNN fine-tuned on the training set with super-resolution}
\label{table:synth_ensemble}
\begin{tabular}{llll}
\hline\noalign{\smallskip}
CNN & Models & Original val set & Val set from MTL\\
 \noalign{\smallskip}
\hline
\noalign{\smallskip}
& Pre-trained & 0.5972 & 0.4124\\
 & Fine-tuned (head) & 0.5824 & 0.3422\\
EfficientNet-B0~\cite{savchenko2021emotions} & Pre-trained + fine-tuned (head) & 0.6400 & 0.4204\\
& Fine-tuned (all) & 0.6718 & 0.3583\\
 & Pre-trained + fine-tuned (all) & \bf 0.6846& 0.4008\\
\hline
& Pre-trained & 0.6094 & 0.4257\\
& Fine-tuned (head) & 0.6198 & 0.4359\\
MT-EmotiEffNet  & Pre-trained + fine-tuned (head)& 0.6450 &  \bf 0.4524\\
& Fine-tuned (all) & 0.6729 & 0.3654\\
 & Pre-trained + fine-tuned (all) & 0.6818 & 0.3970 \\
\hline
\end{tabular}
\end{center}
\end{table}
\setlength{\tabcolsep}{1.4pt}

According to these results, the original validation set seems to be not representative. For example, fine-tuning all weights leads to an increase in the validation F1 score by 6-9\%. However, it \textit{decreases} the performance by 6\% for the new validation set if facial expressions from other subjects should be predicted. It is important to highlight that the pre-trained models are characterized by high F1 scores (41-42\%) in the latter case. The proposed MT-EmotiEffNet is still 1\% more accurate than EfficientNet-B0~\cite{savchenko2021emotions}. What is more important, the fine-tuning of the new head of our model increased performance by 1\%, while the same fine-tuning of EfficientNet-B0 caused significant degradation of the FER quality. The best results are achieved by blending the predictions of a pre-trained model and its fine-tuned head. In fact, this blending can be considered as adding a new classifier head to visual embeddings extracted by MT-EmotiEffNet, so only one inference in a deep CNN is required. For sure, the best results on the original validation set are provided by an ensemble with the model with re-training of all weights. The difference in confusion matrices of pre-trained MT-EmotiEffNet and our blending is shown in Fig.~\ref{fig:3}. Hence, our final submission included both ensembles that showed the best F1 score on the original validation set and the new one from the MTL challenge.

\begin{figure}[t]
\centering
\begin{subfigure}{.41\textwidth}
 \centering
 \includegraphics[width=\linewidth]{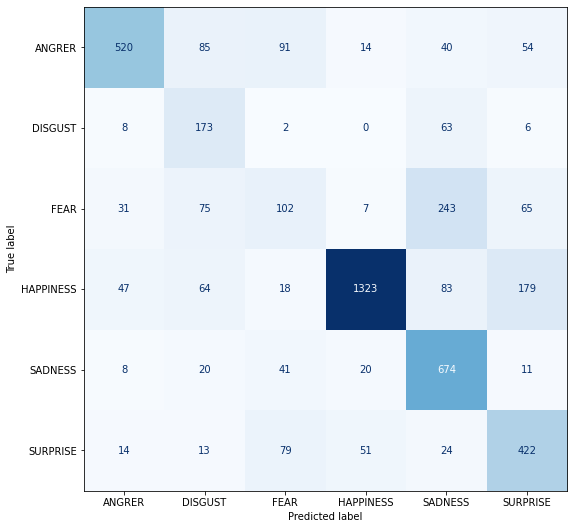}
 \caption{}
\end{subfigure}
\begin{subfigure}{.41\textwidth}
 \centering
 \includegraphics[width=\linewidth]{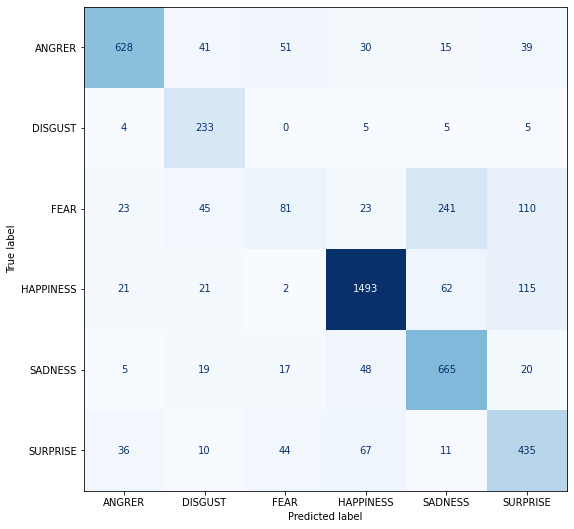}
 \caption{}
\end{subfigure}
 \caption{Confusion matrix for learning from synthetic data: (a) pre-trained MT-EmotiEffNet; (b) ensemble of pre-trained and fine-tuned MT-EmotiEffNets.}
 \label{fig:3}
 \end{figure}

\ifdefined\DEBUG

\begin{figure}[t]
\centering
\begin{subfigure}{.32\textwidth}
 \centering
 \includegraphics[width=\linewidth]{Fig3a}
 \caption{}
\end{subfigure}
\begin{subfigure}{.32\textwidth}
 \centering
 \includegraphics[width=\linewidth]{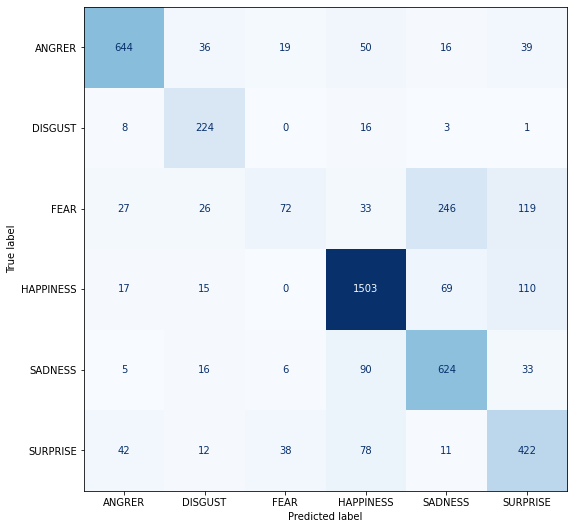}
 \caption{}
\end{subfigure}
\begin{subfigure}{.32\textwidth}
 \centering
 \includegraphics[width=\linewidth]{Fig3c}
 \caption{}
\end{subfigure}
 \caption{Confusion matrix for learning from synthetic data: (a) pre-trained MT-EmotiEffNet; (b) fine-tuned MT-EmotiEffNet; (c) ensemble of pre-trained and fine-tuned MT-EmotiEffNets.}
 \label{fig:3}
 \end{figure}

COMPLETE INFO!!!!!!!!!!!!!!!!!
\setlength{\tabcolsep}{4pt}
\begin{table}
\begin{center}
\caption{Ablation study for the learning from synthetic data challenge}
\label{table:synth_ablation}
\begin{tabular}{p{0.22\linewidth}llllll}
\hline\noalign{\smallskip}
  & \multicolumn{2}{l}{Pre-trained} & \multicolumn{2}{l}{Fine-tuned (orig)} & \multicolumn{2}{l}{Fine-tuned (SR)}\\
 Model & F1 score & Accuracy & F1 score & Accuracy & F1 score & Accuracy \\
 \noalign{\smallskip}
\hline
\noalign{\smallskip}
 EfficientNet-B0~\cite{savchenko2021emotions} &  0.5972 & 0.6659 & 0.6490 & 0.7332 & 0.6718 & 0.7383 \\
 EfficientNet-B2~\cite{savchenko2022cvprw} & 0.5206 & 0.6015 & 0.6412 & 0.7308 & 0.6544 & 0.7385\\ 
MT-EmotiEffNet & 0.6094 & 0.6882 & 0.6324 &  0.7268 & 0.6729 & 0.7471\\
\hline
\end{tabular}
\end{center}
\end{table}
\setlength{\tabcolsep}{1.4pt}

Table~\ref{table:synth_ensemble}.

\setlength{\tabcolsep}{4pt}
\begin{table}
\begin{center}
\caption{Results of an ensemble of pre-trained and fine-tuned CNNs for the learning from synthetic data challenge}
\label{table:synth_ensemble}
\begin{tabular}{p{0.18\linewidth}p{0.2\linewidth}llll}
\hline\noalign{\smallskip}
  & & \multicolumn{2}{l}{Original validation set }& \multicolumn{2}{l}{Validation set from MTL}\\
 CNN & Models & F1 score & Accuracy & F1 score & Accuracy \\
 \noalign{\smallskip}
\hline
\noalign{\smallskip}
& Pre-trained & 0.5972 & 0.6659 & \\
 & Fine-tuned (orig) & 0.6467 & 0.7407 & \\
EfficientNet-B0~\cite{savchenko2021emotions} & Pre-trained + fine-tuned (orig) & 0.6742 & 0.7578 & \\
& Fine-tuned (SR) & 0.6718 & 0.7383 & \\
 & Pre-trained + fine-tuned (SR) & 0.6846 & 0.7486& \\
\hline
& Pre-trained & 0.6094 & 0.6882 & \\
 & Fine-tuned (orig) & 0.6324 & 0.7268 & \\
MT-EmotiEffNet & Pre-trained + fine-tuned (orig) & 0.6676 & 0.7525 & \\
& Fine-tuned (SR) & 0.6729 & 0.7471 & \\
 & Pre-trained + fine-tuned (SR) & 0.6818 & 0.7570 & \\
\hline
\end{tabular}
\end{center}
\end{table}
\setlength{\tabcolsep}{1.4pt}

\fi

\section{Conclusions}\label{sec:4}

In this paper, we presented the multi-task EfficientNet-based model for simultaneous recognition of facial expressions, valence, and arousal that was pre-trained on static photos from the AffectNet dataset. Based on this model, we introduced two novel pipelines for MTL (Fig.~\ref{fig:1}) and LSD (Fig.~\ref{fig:2}) challenges in the fourth ABAW competition~\cite{kollias2022abaw}. It was experimentally demonstrated that the proposed model significantly improves the results of either baseline VGGFACE CNN~\cite{kollias2022abaw4} or single-task EfficientNet-B0~\cite{savchenko2021emotions} for both tasks (Tables~\ref{table:mtl},~\ref{table:synth_ensemble}). It is worth noting that the best performance in both challenges is obtained by original pre-trained MT-EmotiEffNet without a need for fine-tuning of all weights on Aff-Wild2 data. Thus, the well-known issue of affect analysis techniques, namely, the subgroup distribution shift, is partially overcome in our models by training simple MLP-based predictors on top of facial emotional features extracted by the MT-EmotiEffNet.



\section*{Acknowledgements} The research is supported by RSF (Russian Science Foundation) grant 20-71-10010. The work in Section~\ref{sec:3} was implemented in the framework of the Basic Research Program at the National Research University Higher School of Economics (HSE University), Russia in 2022.

%
%
\bibliographystyle{splncs04}
\bibliography{eccv2022submission}
\end{document}